\documentclass[letterpaper]{article} 
\usepackage{aaai24}  
\usepackage{times}  
\usepackage{helvet}  
\usepackage{courier}  
\usepackage[hyphens]{url}  
\usepackage{graphicx} 
\usepackage{natbib}  
\usepackage{caption} 
\usepackage{algorithm}
\usepackage{algorithmic}
\usepackage{multirow}

\usepackage{newfloat}
\usepackage{listings}
\DeclareCaptionStyle{ruled}{labelfont=normalfont,labelsep=colon,strut=off} 
\floatstyle{ruled}
\newfloat{listing}{tb}{lst}{}
\floatname{listing}{Listing}
\newcommand\blfootnote[1]{%
  \begingroup
  \renewcommand\thefootnote{}\footnote{#1}%
  \addtocounter{footnote}{-1}%
  \endgroup
}

\title{Bio-SIEVE: Exploring Instruction Tuning Large Language Models for Systematic Review Automation}
\author{Ambrose Robinson,\textsuperscript{\rm 1} William Thorne,\textsuperscript{\rm 1} Ben Wu,\textsuperscript{\rm 1} Abdullah Pandor,\textsuperscript{\rm 2} Munira Essat,\textsuperscript{\rm 2} Mark Stevenson,\textsuperscript{\rm 1} Xingyi Song\textsuperscript{\rm 1}}
\affiliations{
  Department of Computer Science\textsuperscript{\rm 1}/School of Health and Related Research\textsuperscript{\rm 2}, The University of Sheffield \\
  ambroser53@gmail.com, \{arobinson10, wthorne1, bpwu1, a.pandor, m.essat, mark.stevenson, x.song\}@sheffield.ac.uk
}

\begin{document}
\maketitle
\begin{abstract}

Medical systematic reviews can be very costly and resource intensive. We explore how Large Language Models (LLMs) can support and be trained to perform literature screening when provided with a detailed set of selection criteria. Specifically, we instruction tune LLaMA and Guanaco models to perform abstract screening for medical systematic reviews. Our best model, Bio-SIEVE, outperforms both ChatGPT and trained traditional approaches, and generalises better across medical domains. However, there remains the challenge of adapting the model to safety-first scenarios. We also explore the impact of multi-task training with Bio-SIEVE-Multi, including tasks such as PICO extraction and exclusion reasoning, but find that it is unable to match single-task Bio-SIEVE's performance. We see Bio-SIEVE as an important step towards specialising LLMs for the biomedical systematic review process and explore its future developmental opportunities. We release our models, code and a list of DOIs to reconstruct our dataset for reproducibility.\blfootnote{Preprint. Under review.}

\end{abstract}

\section{Introduction}

Systematic reviews (SR) are widely used in fields such as medicine, public health and software engineering where they help to ensure that decisions are based on the best available evidence. However, they are time-consuming and expensive to create. Expensive specialist time must be spent evaluating natural language documents. This is becoming infeasible due to the exponentially increasing release of literature, especially in the biomedical domain \cite{ zhaoRecentAdvancesBiomedical2021}. \citet{michelsonSignificantCostSystematic2019} estimated that the average SR cost \$141,194 and takes a single scientist an average of 1.72 years to complete.

Automation approaches have been introduced to assist in alleviating these issues, targeting different stages of the process. The most targeted stages are searching, screening and data extraction. It is standard practice for screening solutions to utilise an active learning approach. A human is "in the loop" labelling the model's least certain samples and ranking articles by relevance \cite{sadriContinuousActiveLearning2022, wallaceDeployingInteractiveMachine, wangNeuralRankersEffective2022}. However, stopping criteria is a common insufficiency, often being left to the end user and risking missed relevancy. Regardless, this does not lead to an out-the-box solution and requires significant screening before satisfactory performance is achieved \cite{przybylaPrioritisingReferencesSystematic2018}.

Language models like BERT \cite{devlinBERT} and T5 \cite{RaffelT5} have been applied to screening prioritisation \cite{sadriContinuousActiveLearning2022, yangGoldilocksJustRightTuning2022, wangNeuralRankersEffective2022} and classification \cite{moreno-garciaNovelApplicationMachine2023, qinNaturalLanguageProcessing2021}. However, model input size has been a common limitation and zero-shot performance severely lacked compared to basic trained models like Support Vector Machines (SVM) or traditional methods such as Query Likelihood Modelling (QLM). Given the general-purpose capability of LLMs like GPT-3.5-turbo (hence referred to as ChatGPT), studies have attempted to evaluate the ability to assist in screening classification using a zero-shot approach with promising yet varied results, evoking the need for a specialised solution. 

Reviewers must also provide reasons for excluding potentially relevant articles. Automating this task could reduce workload as a qualitative filtering mechanism - where sensitivity (recall) is essential, excluded reviews could be briefly inspected to validate their exclusion. Exclusion reasons can also provide reviewers using these tools with an insight into the model's decision process.

Our contribution is a family of instruction fine-tuned Large Language Models, Bio-SIEVE (Biomedical Systematic Include/Exclude reViewer with Explanations), that attempts to assist in the SR process via classification. By incorporating the existing and expansive Cochrane Review knowledge base via instruction tuning, Bio-SIEVE establishes a strong baseline for inclusion or exclusion classification screening of potential eligible studies given their abstract for unseen SRs. Bio-SIEVE is highly flexible and able to consider specific details of a review's objectives and selection criteria without the need to retrain.

The task we explore is more challenging than existing work as it requires filtering of more subtly irrelevant articles. Previous work has mainly comprised of screening for simple topics or single selection criterion \cite{syrianiAssessingAbilityChatGPT, moreno-garciaNovelApplicationMachine2023}. We filter by an arbitrary set of selection criteria and objectives and extend this problem by introducing the novel, challenging task of exclusion reasoning.

We investigate the efficacy of different instruction tuning methods on our data with an ablation study. Following the work of \citet{vuExploringPredictingTransferability2020,sanhMultitaskPromptedTraining2022}, we train a set of models on the multi-task paradigm of PICO extraction and exclusion justification in an attempt to leverage beneficial cross-task transfer. As \citet{longpreFlanCollectionDesigning2023} found that treating generalised instruction tuning as pretraining led to pareto improvements, we fine-tune on top of Guanaco in addition to LLaMA. We find that multi-task transfer is limited but instruction tuned pretraining caused marginal improvements. We also find that training on our dataset leads to highly accurate exclusion of inappropriate studies, e.g. excluding muscle trauma studies from oral health reviews. Finally, Bio-SIEVE-Multi shows promise for the task of inclusion reasoning but fails to match the performance of ChatGPT in preference rankings.

We believe that Bio-SIEVE lays the foundation for LLMs specialised for the SR process, paving the way for future developments for generative approaches to SR automation. We open-source our codebase\footnote{\url{https://github.com/ambroser53/Bio-SIEVE}} and the means with which to recreate our datasets. We also release our adapter weights on HuggingFace\footnote{\url{https://huggingface.co/Ambroser53/Bio-SIEVE}} for reuse and further development.

\section{The Systematic Review Process}

\begin{figure}[t]
\centering
\includegraphics[width=0.9\columnwidth]{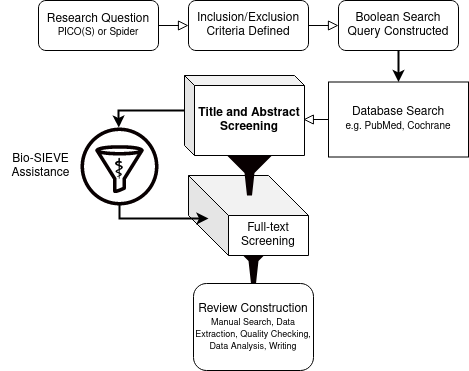} 
    \caption{A simple representation of the systematic review process depicting the stage which Bio-SIEVE aims to assist. The black funnels are the monotonous and highly resource intensive bottlenecks of the process.}
    \label{figure:sysrevprocess}
\end{figure}

The systematic review process is a series of steps mapping a comprehensive plan for the study of a specific research field. This results in an effective summarisation of research material in a particular area or to answer a particular question within a domain. 

Initially the reviewer establishes a research question from which a selection criteria is developed that defines the scope of the project and therein the criteria for study relevance. The Population, Intervention, Comparison, Outcome (PICO) framework is a tool that can be used to define the parameters of a study. Other frameworks also exist such as PICOS and SPIDER \cite{methleyPICOPICOSSPIDER2014, Cates134, klodaComparisonPatientIntervention2020}. This, along with a preliminary search, helps to establish the reviews inclusion and exclusion criteria.

Once the parameters of the study are sufficiently defined, a Boolean query is constructed for use in the searching of large databases in order maximise the recall of as many relevant articles as possible and is refined in an iterative process \cite{wangCanChatGPTWrite2023}. In the next stage, the relevance of each study to the review is assessed via evaluation of the study's title and abstract. The recall from Boolean queries can lead to massive amounts of documents and the time and cost of this stage can be further exacerbated by "double-screening" and "safety-first" approaches that require multiple reviewers independently carrying out the same relevance screening \cite{shemiltUseCosteffectivenessAnalysis2016}. The following stage is full-text screening where it is hoped that the majority of irrelevant studies have been discarded since, compared to title and abstract, obtaining the full-text of studies is not necessarily trivial \cite{tawfikStepStepGuide2019}.

The final stages consist of: adding included reviews based on manual searching; data extraction of relevant info and quality checking; data checking and double checking; analysis, and writing.

As depicted in Figure \ref{figure:sysrevprocess}, Bio-SIEVE targets the title and abstract screening phase given the objectives and selection criteria of the study established by the review team earlier in the process and the abstract of the study being screened. This phase is the most appropriate given the current capability of LLMs as full-text screening requires longer context lengths.

\section{Related Work}
There have been a number of approaches to automating the SR process. These works are delineated into \textit{classification} techniques, which provide a distinct inclusion or exclusion label, and \textit{prioritisation} techniques, which assist in screening by ranking reviews by relevance. Where classifiers aim to directly reduce the number of manually screened studies, ranking strategies aim to allow the reviewer to stop screening early by considering the top-k returned documents.

Basic screening techniques have matured, for example \citet{marshallMachineLearningIdentifying2018} and \citet{wallaceIdentifyingReportsRandomized2017}, which are n-gram classifiers for randomised control trials, with the latter being integrated into Cochrane Reviews' Evidence Pipeline \cite{HowCochraneUsing}. These methods excel at single, easily generalisable tasks but far more difficult is evaluating articles based on topic and review-specific inclusion criteria. 

Other early classifier techniques utilise ensemble SVMs \cite{wallaceSemiautomatedScreeningBiomedical2010}, Random Forest (RF) \cite{khabsaLearningIdentifyRelevant2016} or Latent Dirichlet Allocation \cite{miwaReducingSystematicReview2014} algorithms with active learning strategies to combat the heavy "exclude" class imbalance that naturally occurs. Many more recent approaches such as Abstrackr \cite{wallaceDeployingInteractiveMachine}, Rayyan \cite{olofssonCanAbstractScreening2017} and RobotAnalyst \cite{przybylaPrioritisingReferencesSystematic2018} simply take this regime and streamline its usability. However, there are some clear issues with this approach. For example, \citet{przybylaPrioritisingReferencesSystematic2018} found that RobotAnalyst required anywhere between 29.26\% to 93.11\% of their study collection pool to be manually screened before 95\% recall relevance was achieved.

\citet{qinNaturalLanguageProcessing2021} was first to apply transformers to classification in the context of SRs, yet found fine-tuned BERT was outperformed by their Light Gradient Boosting Machine. Active learning with transformers was applied to Technology Assisted Review tasks \cite{sadriContinuousActiveLearning2022} with \citet{yangGoldilocksJustRightTuning2022} finding that the amount of pretraining before active learning is crucial. \citet{wangNeuralRankersEffective2022} evaluated a variety of BERT models for relevance ranking both fine-tuned and zero-shot, but disregarded the models zero-shot capability after poor results. Most recently, \citet{moreno-garciaNovelApplicationMachine2023} applied BART "zero-shot" with input embeddings on sets of abstracts queried with short questions over a specific selection criterion but saw poor performance unless combined with an RF or SVM.

The recent widespread adoption of ChatGPT has invigorated attempts to utilise LMs for classification, especially in the medical domain. \citet{qureshiAreChatGPTLarge2023} comments that ChatGPT selected articles when used for relevance screening "could serve as a starting point for refinement depending on the complexity of the question". \citet{wangCanChatGPTWrite2023} quantitatively explored ChatGPTs ability to assist in the searching process by constructing Boolean queries but found that although precision was promising, recall was disappointing and variability from minor prompt changes and even same prompt use brought the reproducibility of its use into question. Methodical studies evaluating ChatGPT's effectiveness in classification have started to emerge. \citet{guoAutomatedPaperScreening2023} reported 91\% exclusion recall but only 76\% recall of included articles when screening a dataset of 24k+ total abstracts where only 538 were inclusion samples. They also remarked on ChatGPT's ability to generate reasoning for exclusions and it's potential for improving SR screening quality. \citet{syrianiAssessingAbilityChatGPT} placed a strong emphasis on reproducibility, setting a temperature of zero to ensure a higher level of consistency and found that, when prompted to be more lenient with inclusions, ChatGPT could be more conservative and sustain high recall of eligible examples given the general topic of the study and the abstract of the potential reference. They concluded that ChatGPT is a viable option. 

We argue the use of ChatGPT unavoidably compromises reproducibility. The alteration and retraining of ChatGPT over time is opaque as \citet{chenHowChatGPTBehavior2023} found that its performance on certain tasks had changed dramatically between March and June of 2023. Furthermore, ChatGPT's size and consumption costs are similarly opaque but, as a generalised model, it can be assumed to be larger than any specialised approach. This elicits the demand for a smaller language model specialised for this task, where the exact model can be referenced and computational resources disclosed.


LLaMA \cite{touvron2023llama} has become a popular foundational model for causal generation as it was made open for non-commercial use in contrast to the GPT family \cite{brown2020language, openai2023gpt4} which has been closed-source since GPT3. Reinforcement-Learning with Human-Feedback (RLHF) \cite{christiano2017deep} has become a popular technique for controlling generated outputs from language models. InstructGPT \cite{instructgpt} applied RLHF to improve the response quality of LLMs and was expanded upon to create ChatGPT which has become the benchmark for zero-shot performance.

Since ChatGPT, many open-source instruction tuned LLMs have emerged to try to match its performance. Guanaco \cite{dettmers_qlora_2023} is a family of LLaMA-based LLMs trained using 4-bit quantization and LoRA. The zero-shot performance of Guanaco-65B on the Vicuna benchmark \cite{vicuna2023} achieves 99.3\% the performance of ChatGPT.

Instruction tuning is a method of fine-tuning where tasks are phrased as natural language prompts and has been shown to improve LLM performance on zero-shot tasks. \cite{weiFINETUNEDLANGUAGEMODELS2022} The detailed ablation study carried out by \citet{longpreFlanCollectionDesigning2023} found that treating instruction tuning as pretraining before downstream task fine-tuning caused faster convergence and often provided better performance overall. \citet{vuExploringPredictingTransferability2020} found that transfer learning with multiple tasks in the same domain could improve the performance of the tasks individually.

Full fine-tuning of LLMs is prohibitively expensive to non-commercial entities; as such, techniques have been developed to minimise computational requirements and training time while maintaining high performance. Based on the hypothesis that parameter updates have a low intrinsic rank \cite{aghajanyan2020intrinsic}, Low Rank Adaptation (LoRA) \cite{hu2021lora} applies a rank decomposition of specified weight matrices while freezing the original network to reduce the trainable parameter count whilst delivering comparable performance to full-finetuning. Combined with 8- or even 4-bit quantization, it is possible to fine-tune 65B parameter models on a single 48GB GPU \cite{dettmers20228bit, dettmers_qlora_2023}. 

\section{Methods}

\textbf{Instruct Cochrane Dataset} We gathered a total of 7,330 medical SRs from all possible topic areas available on the Cochrane Reviews\footnote{\url{www.cochranelibrary.com}} website. Each review contained the objectives and selection criteria along with all considered studies and whether they were included or excluded from the review. Excluded studies were accompanied by a reason for exclusion. Out of these 7,330 reviews were derived a training split of 6,963 and an evaluation split of 367. Each study is treated as an individual data point. The distributions of the separate splits are displayed in Table \ref{table:dataset_stats}.

\begin{figure}[t]
\centering
\includegraphics[width=1\columnwidth]{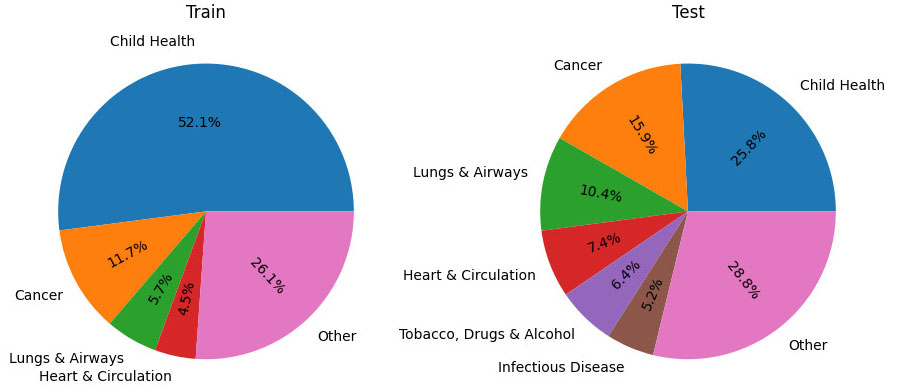} 
\caption{The topic distribution of the inclusion/exclusion classification samples in the train and test splits of the Instruct Cochrane dataset.}
\label{topic_dist_pie}
\end{figure}

Cochrane was selected for review gathering as the review format is standardised. The delineated objectives and selection criteria were suitably informative for review specification and the exhaustive references were clearly categorised into included and excluded. In addition, reviews provided justification for exclusion and descriptions of the population, intervention and outcome of considered reviews created a basis for a multi-task dataset. Comparison data was difficult to retrieve and were not included thus these tasks will hence be referred to as PIO extraction tasks. Topic distribution of the train and test sets can be found in Figure \ref{topic_dist_pie}.

Due to size of the test split and the fact that we are evaluating many models on many different test sets, we instead chose to use a truncated subset of the full test split that maximised the diversity of topics. This allowed the test set to remain the basis of evaluation for the model's generalisation across topics.

\begin{table}[]
\centering
\begin{tabular}{ cccccc }
\hline
Task & Train & Test & Subset & S-1\textsuperscript{st} & Irre. \\
\hline
Inclusion & 43,221 & 576 & 784 & 79 & -\\
Exclusion & 44,879 & 425 & 927 & 29 & 780\\
\hline
Inc/Exc & 88,100 & 1,001 & 1,711 & 108 & 780\\
\hline
Population & 15,501 & - & - & - & -\\
Intervention & 15,386 & - & - & - & -\\
Outcome & 15,518 & - & - & - & -\\
Exc. Reason & 11,204 & - & - & - & -\\
\hline
\textbf{Total} & \textbf{168,842} & \textbf{1,001} & \textbf{1,711} & \textbf{108} & \textbf{780}\\
\hline
\end{tabular}
\caption{Number of samples in each split of the Instruct Cochrane dataset.}
\label{table:dataset_stats}
\end{table}

\textbf{Instruction Tuning Method} Following the work of \citet{chungScalingInstructionFinetunedLanguage2022}, we utilised instruction fine-tuning in order to bolster the efficacy of the fine-tuning process. Input data was formatted with natural language instructions for the tasks of inclusion or exclusion classification, PICO extraction, and exclusion reason generation. To minimise information loss from truncation, the inputted sections were tokenised and scaled down proportional to their length until they fit within the max input token length of 2048. We utilise the Alpaca instruction format \cite{alpaca} with all our models to match Guanaco and to maintain consistency between the Guanaco and base LLaMA models. (See Figure \ref{figure:training_sample} for an example instruction). Full details on our preprocessing methods can be found in Appendix~\ref{appendix:preprocessing}.

\textbf{Safety-First Test Set} Since samples could potentially have been excluded during full-text screening with information unavailable in the abstract, the test split labels may not reflect the appropriate decisions at an abstract screening stage. We curated a safety-first test set by manually annotating include/exclude decisions for a small subset of 108 samples from the test split, with each sample consisting of the objectives and selection criteria for the review and the abstract of the prospective study. The resulting safety-first set was biased toward include with 79 samples and 29 samples treated as exclude. This results in a benchmark that rewards cautious models with high include recall.

Annotation was performed by professional medical systematic reviewers, who were instructed to choose from 3 labels: 'Include', 'Exclude', or 'Insufficient Information'. For evaluation purposes, we treat 'Insufficient Information' as 'Include', since these would be samples that should proceed to full-text screening phase, in keeping with a safety-first approach.

Overall, there were 11 disagreements between our labels and the labels provided by the original reviewers. Our annotators labelled 3 samples as "Include" and 8 samples as "Insufficient Information". In the Instruct Cochrane dataset, these samples were all labelled, after full-text screening, as \textit{exclude}.

This issue of label mismatch extends to the larger test set and the training set. However, applying this manual re-annotation procedure to training data and more testing data was infeasible due to costs, especially since professional medical reviewers were used as annotators.

\begin{figure}[]
\noindent\fbox{%
    \parbox{\linewidth}{%
        \paragraph{Instruction} Given the abstract, selection criteria and objectives should the study be included or excluded? 
        \paragraph{Input} \underline{Abstract}:  This paper describes the study design, methodological considerations, and baseline characteristics of a clinical trial to determine if intense 48 weeks, twice per week Tai Chi practice can reduce the frequency of falls among older adults transitioning to frailty compared to a wellness education program. ... Secondary outcome measurements include ... 
        
        \underline{Objectives}:  To assess the effects benefits and harms of exercise interventions for preventing falls in older people living in the community. 
        
        \underline{Selection Criteria}:  We included randomised controlled trials RCTs evaluating the effects of any form of exercise as a single intervention on falls in people aged 60 years living in the community. We excluded trials focused on particular conditions, such as stroke.
        \paragraph{Response} \underline{Included}
    }%
}
\caption{Example Instruct Cochrane sample used in instruction tuning. \citet{wolfStudyDesignInvestigate2001} is the study being being evaluated for inclusion in \citet{gillespieInterventionsPreventingFalls2012}.} \label{figure:training_sample}
\end{figure}

\textbf{Review Subsets} In order to compare to classifier techniques trained for a specific review, a subset of reviews with a sufficient number of abstracts is required to train and evaluate with k-fold cross validation. For this we took the 13 reviews from the evaluation set that have over 100 associated abstracts resulting in a total of 1711 individual samples. 

\textbf{Irrelevancy Test Set} To ensure that the model is able to exclude wildly irrelevant submissions, we constructed a evaluation set of selection criteria paired with abstracts from a completely distinct topic. Starting from the 13 reviews in the Review Subset, we paired each review with 5 random abstracts from the other reviews. Since each review in this subset is from a different topic area, each instruction prompt formed from this set is guaranteed to be irrelevant (i.e. should be classified as 'exclude').

\section{Experimental Setup}

\subsection{Bio-SIEVE Model Description}
To evaluate the suitability of multi-task transfer learning \cite{vuExploringPredictingTransferability2020,sanhMultitaskPromptedTraining2022} and instruction tuning as pretraining \cite{longpreFlanCollectionDesigning2023} we conduct an ablation, training four Bio-SIEVE models: single-task/multi-task, Guanaco/LLaMA.

We used QLoRA fine-tuning to train LLaMA7b and Guanaco7B on the Instruct Cochrane Train split. For the multi-task versions, 4 A100 80GB cards were used for 40 hours with an effective batch size of 16, whereas the single task versions were trained in the same setup for 24 hours.

In order to maintain consistency with Guanaco \cite{dettmers_qlora_2023}, we fix the hyperparameters during QLora finetuning: 4-bit double quantisation to the NF-4 datatype, 0.1 LoRA dropout, LoRA alpha of 16, LoRA rank 64, LoRA adapters on all layers but without biases and a learning rate of 2e-4 with no warmup or learning rate decay. LLaMA base model LoRA weights were randomly initialised with seed 0. Each model was trained for 8 epochs and the best model on the safety-first set was selected for comparison.

\subsection{Evaluation Methodology}
We evaluate inclusion and exclusion performance through accuracy for an overview of model understanding, and precision and recall for the inclusion class as a more direct evaluation of the model's real world applicability to SR automation. We compare the different permutations of Bio-SIEVE to a series of zero-shot baselines and trained models using the standard approach on the review subset. Each experiment was ran once with a temperature of 0 and no sampling.

\textbf{Zero-shot comparisons} To establish a baseline for our task, we query three generic, instruction tuned models: ChatGPT, Guanaco7B, and Guanaco13B. We test Guanaco7B as it represents the baseline performance of the Guanaco7B versions of Bio-SIEVE, prior to finetuning. ChatGPT is used as the state-of-the-art comparison and Guanaco13B to observe how model scale affects zero-shot performance. To alter the tasks for evaluation with ChatGPT, we use the prompt designed by \citet{syrianiAssessingAbilityChatGPT}.  We include additional information on the objectives and selection criteria of the SR for additional context since our regular test set is more challenging (See Figure~\ref{figure:gpt_prompt} in Appendix~\ref{appendix:chatgpt_prompt}).

In order to compare against the zero-shot method defined in \citet{wangNeuralRankersEffective2022}, we use Bio-BERT \cite{Lee_2019} finetuned on the MS MARCO dataset\footnote{\url{https://huggingface.co/nboost/pt-biobert-base-msmarco}} (hence Bio-BERT-MSM) \cite{gaoRethinkTrainingBERT2021, MSMARCO} and evaluate performance on our Test, Safety-first and Subset data zero-shot.

\textbf{Inclusion Exclusion Baselines} To simulate the active learning approach standard to the field, we applied 5-fold cross validation using a logistic regression model to determine the average performance across the 13 reviews within the large review subset. We also fine-tuned and evaluated Bio-BERT-MSM in the same manner. 

We applied a standard data pre-processing methodology for logistic regression. Each abstract was lowercased, had stopwords removed, and was lemmatized using NLTK \cite{bird2009natural}. A new tokenizer was trained for each review based on TF.IDF. All training and tokenization was performed using Scikit-Learn \cite{pedregosa2011scikit}. For Bio-BERT-MSM, only the abstract was provided when fine-tuning. However, when evaluating zero-shot performance the review's objectives and selection criteria were also provided utilising the Huggingface Zero-shot Classification Pipeline as in \citet{moreno-garciaNovelApplicationMachine2023}.

\textbf{Exclusion Reasoning} We evaluate the multi-task model variants' exclusion reasons via 5-star ranking. 81 exclusion tasks are taken from 3 reviews out of the 13 review subset. These are selected to best fit our expertise to speed up the process. All samples are validated to ensure that the justification could be derived from only the objectives, selection criteria and abstract. Outputs with significant generation artifacts are penalised 1 star with a minimum score of zero. A rating of 5 represents a perfect match with the original reviewer's justification.

\section{Results}

\begin{table*}[]
\centering
\begin{tabular}{ |l|ccc|ccc|ccc|c| }
\hline
\multicolumn{1}{|c}{} &
    \multicolumn{3}{|c}{\textbf{Test}} &
    \multicolumn{3}{|c}{\textbf{Subset}} &
    \multicolumn{3}{|c}{\textbf{Safety-first}} &
    \multicolumn{1}{|c|}{\textbf{Irre.}}\\
\cline{2-11}
\multicolumn{1}{|c}{\multirow{-2}{*}{\textbf{Model}}} &
    \multicolumn{1}{|c}{\textbf{Pre.}} &
    \multicolumn{1}{|c}{\textbf{Rec.}} &
    \multicolumn{1}{|c}{\textbf{Acc.}} &
    \multicolumn{1}{|c}{\textbf{Pre.}} &
    \multicolumn{1}{|c}{\textbf{Rec.}} &
    \multicolumn{1}{|c}{\textbf{Acc.}} &
    \multicolumn{1}{|c}{\textbf{Pre.}} &
    \multicolumn{1}{|c}{\textbf{Rec.}} &
    \multicolumn{1}{|c}{\textbf{Acc.}} &
    \multicolumn{1}{|c|}{\textbf{Acc.}}\\
\cline{1-11}
\textit{Logistic Regression*} &
    \multicolumn{1}{c|}{-} &
    \multicolumn{1}{c|}{-} &
    \multicolumn{1}{c|}{-} &
    \multicolumn{1}{c|}{0.79} &
    \multicolumn{1}{c|}{0.78} &
    \multicolumn{1}{c|}{0.80} &
    \multicolumn{1}{c|}{-} &
    \multicolumn{1}{c|}{-} &
    \multicolumn{1}{c|}{-} &
    \multicolumn{1}{c|}{-}\\
\textit{Bio-BERT-MSM*} &
    \multicolumn{1}{c|}{-} &
    \multicolumn{1}{c|}{-} &
    \multicolumn{1}{c|}{-} &
    \multicolumn{1}{c|}{0.43} &
    \multicolumn{1}{c|}{0.30} &
    \multicolumn{1}{c|}{0.50} &
    \multicolumn{1}{c|}{-} &
    \multicolumn{1}{c|}{-} &
    \multicolumn{1}{c|}{-} &
    \multicolumn{1}{c|}{-}\\
\cline{1-11}
\textit{ChatGPT \textsuperscript{\textdagger} (ZS)} &
    \multicolumn{1}{c|}{0.59} &
    \multicolumn{1}{c|}{\textbf{0.96}} &
    \multicolumn{1}{c|}{0.60} &
    \multicolumn{1}{c|}{0.50} & %
    \multicolumn{1}{c|}{0.86} &
    \multicolumn{1}{c|}{0.55} & %
    \multicolumn{1}{c|}{0.79} &
    \multicolumn{1}{c|}{0.86} &
    \multicolumn{1}{c|}{\textbf{0.73}} &
    \multicolumn{1}{c|}{0.98}\\
\textit{Guanaco13B (ZS)} &
    \multicolumn{1}{c|}{0.58} &
    \multicolumn{1}{c|}{0.95} &
    \multicolumn{1}{c|}{0.56} &
    \multicolumn{1}{c|}{0.47} & %
    \multicolumn{1}{c|}{\textbf{0.96}} &
    \multicolumn{1}{c|}{0.47} & %
    \multicolumn{1}{c|}{0.71} &
    \multicolumn{1}{c|}{0.90} &
    \multicolumn{1}{c|}{0.67} &
    \multicolumn{1}{c|}{0.03}\\
\textit{Guanaco7B (ZS)} &
    \multicolumn{1}{c|}{0.57} &
    \multicolumn{1}{c|}{0.95} &
    \multicolumn{1}{c|}{0.56} &
    \multicolumn{1}{c|}{0.46} & %
    \multicolumn{1}{c|}{0.95} &
    \multicolumn{1}{c|}{0.46} & %
    \multicolumn{1}{c|}{0.73} &
    \multicolumn{1}{c|}{\textbf{0.96}} &
    \multicolumn{1}{c|}{0.72} &
    \multicolumn{1}{c|}{0.02}\\
\textit{Bio-BERT-MSM (ZS)} &
    \multicolumn{1}{c|}{0.69} &
    \multicolumn{1}{c|}{0.14} &
    \multicolumn{1}{c|}{0.54} &
    \multicolumn{1}{c|}{0.46} & %
    \multicolumn{1}{c|}{0.93} &
    \multicolumn{1}{c|}{0.47} & %
    \multicolumn{1}{c|}{0.66} &
    \multicolumn{1}{c|}{0.93} &
    \multicolumn{1}{c|}{0.64} &
    \multicolumn{1}{c|}{0.09}\\
\cline{1-11}
\textit{LLaMA7B (Single)} &
    \multicolumn{1}{c|}{0.77} &
    \multicolumn{1}{c|}{0.79} &
    \multicolumn{1}{c|}{0.74} &
    \multicolumn{1}{c|}{0.65} & %
    \multicolumn{1}{c|}{0.83} &
    \multicolumn{1}{c|}{0.71} & %
    \multicolumn{1}{c|}{0.88} &
    \multicolumn{1}{c|}{0.72} &
    \multicolumn{1}{c|}{0.72} &
    \multicolumn{1}{c|}{0.96}\\
\textit{LLaMA7B (Multi)} &
    \multicolumn{1}{c|}{0.88} &
    \multicolumn{1}{c|}{0.64} &
    \multicolumn{1}{c|}{0.74} &
    \multicolumn{1}{c|}{\textbf{0.85}} & %
    \multicolumn{1}{c|}{0.70} &
    \multicolumn{1}{c|}{0.80} & %
    \multicolumn{1}{c|}{0.91} &
    \multicolumn{1}{c|}{0.38} &
    \multicolumn{1}{c|}{0.52} &
    \multicolumn{1}{c|}{0.97}\\
\textit{Guanaco7B (Single)} &
    \multicolumn{1}{c|}{0.85} &
    \multicolumn{1}{c|}{0.82} &
    \multicolumn{1}{c|}{\textbf{0.82}} &
    \multicolumn{1}{c|}{0.76} & %
    \multicolumn{1}{c|}{0.85} &
    \multicolumn{1}{c|}{\textbf{0.81}} & %
    \multicolumn{1}{c|}{0.88} &
    \multicolumn{1}{c|}{0.62} &
    \multicolumn{1}{c|}{0.66} &
    \multicolumn{1}{c|}{0.98}\\
\textit{Guanaco7B (Multi)} &
    \multicolumn{1}{c|}{\textbf{0.90}} &
    \multicolumn{1}{c|}{0.64} &
    \multicolumn{1}{c|}{0.75} &
    \multicolumn{1}{c|}{0.81} & %
    \multicolumn{1}{c|}{0.69} &
    \multicolumn{1}{c|}{0.78} & %
    \multicolumn{1}{c|}{\textbf{0.95}} &
    \multicolumn{1}{c|}{0.46} &
    \multicolumn{1}{c|}{0.58} &
    \multicolumn{1}{c|}{\textbf{0.99}} \\
\cline{1-11}
\end{tabular}
\caption{Results of inclusion/exclusion classification for a logistic regression baseline, the zero-shot (L)LMs and the Bio-SIEVE variants. Test, Subset and Safety-first metrics are precision, recall and accuracy. Single variants were only trained on the task of include/exclude classification. Multi models were also trained on PIO extraction and exclusion reasoning. \textbf{*} indicates results when trained/fine-tuned with 5-fold cross validation on the Review Subset.
\textsuperscript{\textdagger}ChatGPT version as of 27/07/23}
\label{table:classification_results}
\end{table*}

\begin{figure}[h]
\centering
\includegraphics[width=1\columnwidth]{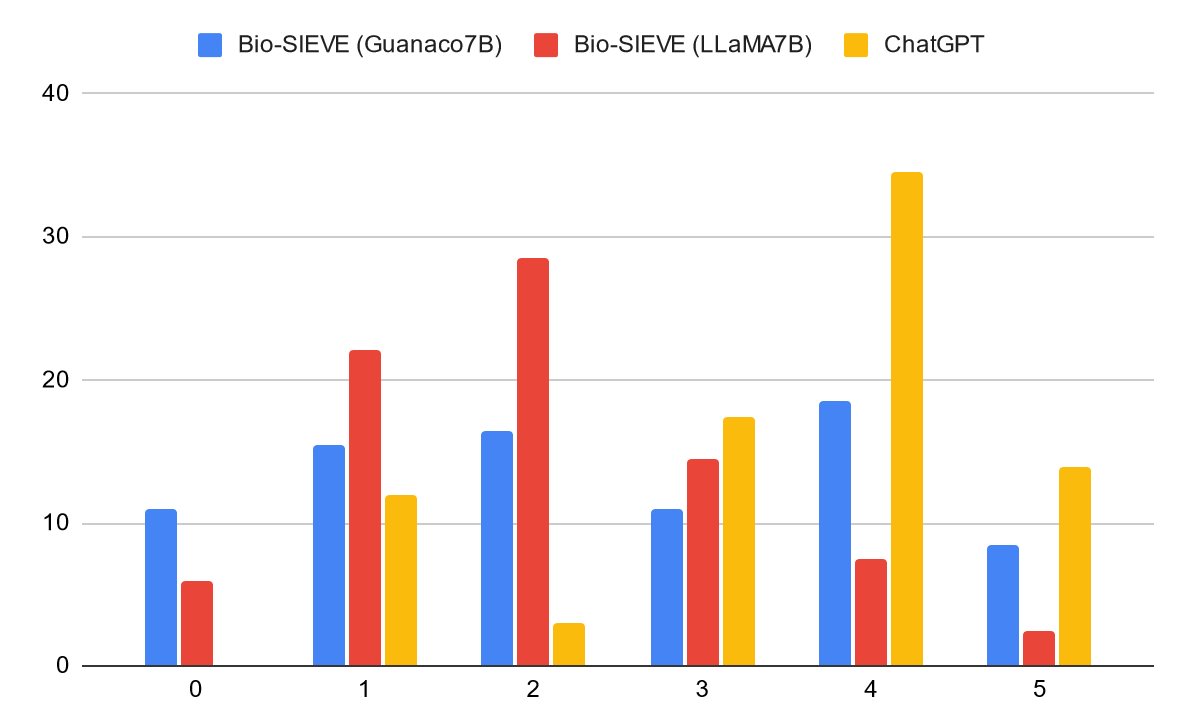}
\caption{Statistics for model generated exclusion reasons when scored by two experts independently given the original author's exclusion justification.}
\label{fig:exclusion_results}
\end{figure}

Results for the classification task for each dataset is presented in Table \ref{table:classification_results}. Preference results for exclusion reason generation ranking is provided in Figure \ref{fig:exclusion_results}. Agreement via Pearson’s correlation coefficient between our two independent experts was r=.84 for our Guanaco7B variant, r=.42 for our LLaMA7b variant and r=.62 for ChatGPT generations and p<.001.

\textbf{Our trained models achieve better accuracy scores than ChatGPT on the Test and Subset Eval sets} Our best performing model, Guanaco7B (Single), achieved 0.82 accuracy on the Test Set. By comparison, ChatGPT achieved 0.6 accuracy. Similarly, for the Subset, Guanaco7b (Single) achieved accuracy 0.26 higher than ChatGPT whilst only reducing inclusion recall by 0.01.

\textbf{Our trained models slightly outperform active-learning style models specialised for a single review} Guanaco7B (Single) achieves 0.81 accuracy on the Subset eval set. The next best model is the Logistic Regression Baseline, which achieved 0.8. We highlight that this LR baseline had a data advantage over our generalised models: separate Logistic Regression models were trained for each individual review in the Subset whereas our trained models relied on only one single fine-tuned LLM for entire Subset. Additionally, the logistic regression models used 80\% of samples per Subset review for training data (5 fold cross-validation). By contrast, our trained models had never seen any of the reviews or studies in the Subset Evaluation set during training. 

\textbf{ChatGPT is able to be more lenient, allowing it to perform well on the safety-first dataset} ChatGPT tended to include reviews rather than exclude, leading to high recall at the expense of precision. (Note: this was due to an explicit prompt instruction to 'be lenient') This means that it performed strongly on the safety-first set with an  accuracy of 0.73. Our best model on this subset, LLaMA7B Single achieved higher precision (0.88) but slightly lower accuracy (0.72). We did not experiment with leniency prompting or thresholding for our trained models but anticipate that this would further improve results.

\textbf{Single-task training was more effective than multi-task training} The single-task models tended to be include recall oriented and higher performing with the Guanaco7B-Single variant outperforming all our other models by at least 0.07 accuracy whilst preserving the highest inclusion recall of 0.82 and 0.85 on Test and Subset respectively.

\textbf{Irrelevancy test: open-source LLMs must be fine-tuned in order to be effective systematic reviewers} Both ChatGPT and our trained models perform very well on the irrelevancy test, demonstrating that these models are able to effectively exclude off-topic abstracts. However, other zero-shot models perform very poorly (Guanaco7B achieves only 0.02), revealing that these open-source models are unsuitable for use in the zero-shot setting as they include many highly irrelevant abstracts.

\textbf{ChatGPT provides the best exclusion reasons} ChatGPT managed an average score of 3.4 in our rankings in comparison to 2.4 and 2.0 for the Guanaco7B and LLaMA7B Bio-SIEVE-Multi variants and we therefore treat ChatGPT as the current state-of-the-art for this exclusion reasoning task. Bio-SIEVE-Multi variants managed to match the quality of ChatGPT for 45\% of samples but there were minimal examples of Bio-SIEVE exceeding its quality (6-7\%). Overall, the quality of exclusion reasons remains poor: ChatGPT generated subpar or incorrect reasons for 83\% of samples.

\begin{figure*}[t]
\centering
\includegraphics[width=1\textwidth]{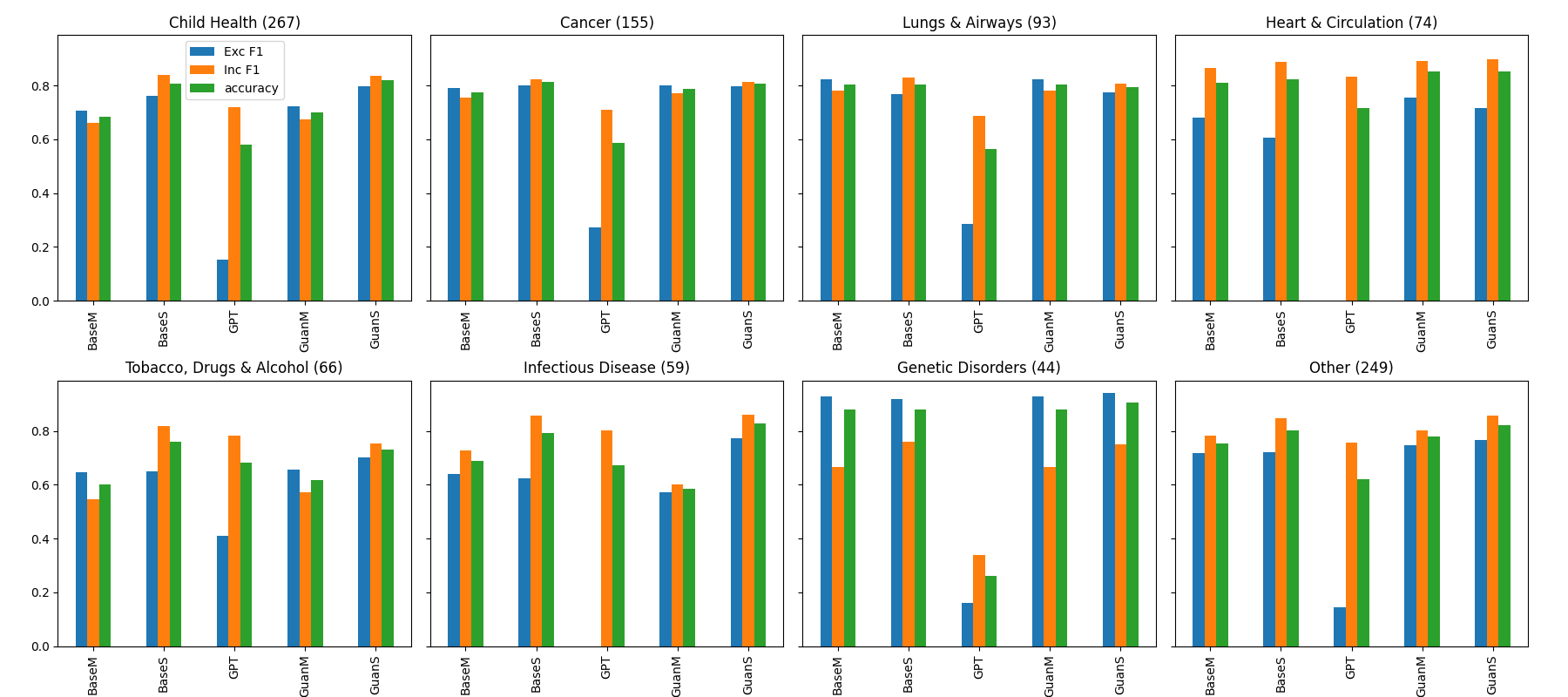} 
\caption{Performance measured by F1-score for each metric for all models compared to ChatGPT on different medical domain topics within the test set. ChatGPT excluded no samples for topics where no exclude bar is present.}
\label{fig:topic_results}
\end{figure*}

\section{Discussion}

\textbf{Model Comparison}
Bio-SIEVE takes a more balanced approach to classification as shown by its consistently higher precision but relatively small decrease in recall. This suggests a greater ability to reason over the selection criteria. In contrast, ChatGPT, using the prompt format of \cite{syrianiAssessingAbilityChatGPT}, tends to be overly lenient and overly inclusive. This is demonstrated when evaluating performance broken down per topic, as shown in Figure \ref{fig:topic_results}. 

Our models perform consistently across review topics whereas ChatGPT performance varies greatly. For "Genetic Disorders" ChatGPT results are significantly below other models; for "Heart \& Circulation" and "Infectious Disease" topics, it predicted "Include" for all samples and never "Exclude". This draws into question the extent of ChatGPT's generality and underscores the necessity to assess model blind spots prior to their endorsement for real world application.

Other zero shot model suffer this problem to an even greater extent. They tend to include everything, even abstracts from unrelated topics, as shown in the Irrelevancy column of \ref{fig:exclusion_results}. Therefore, despite high include recall performance on all three evaluation sets, they do not serve any practical benefit to reviewers. This shows that, despite claims of performance matching ChatGPT on chatbot benchmarks like Vicuna \cite{vicuna2023}, open-source models (when they are not fine-tuned to a task) are still a long way off reaching similar zero-shot capabilities.

Bio-SIEVE outperforming the active learning style Logistic Regression models is also an achievement. This validates that LLMs can facilitate reasoning of SR criteria with language, a far more accessible and cost effective method of automation when compared to the training of models that learn the criterion of individual SRs which has been the de facto approach for over a decade.

\textbf{Effect of Training Data} Training data topic imbalance had no noticeable effect on generalisation. For example,over half the training samples were in the "Child Health" topic yet Bio-SIEVE obtained similar if not better results on "Heart \& Circulation" samples, which made up only 4.5\% of the training data. "Genetic Disorder" topic performance is strong despite only making up 1.2\% of the training data.

We generally find that for multi-task models, cross-task transfer between PIO, Exclusion reasoning and Include/Exclude classification harms performance. We speculate two reasons for this 1) Dataset Imbalance: PIO data was only available for included studies which may have resulted in tasks concentrated around a smaller subset of topics with reduced variation 2) Hallucinations - exclusion reasons and PIO information often relied on information from full-text screening. Training the model to extract this information from the abstract when not present may have encouraged it to hallucinate and/or overfit to the training data.

\section{Conclusion and Further Work}
In this paper, we demonstrate the effectiveness of training open-source LLMs to perform biomedical title/abstract screening. Our trained models achieve significant accuracy improvements over ChatGPT, and are specialised for the healthcare domain. Our results also reveal the dangers of relying on ChatGPT's zero-shot performance: its accuracy is very uneven across healthcare topics, performing particularly poorly on Genetic Disorders and being excessively lenient for Infectious Disease and Heart \& Circulation. This is in addition to reproducibility concerns arising from the opacity of closed-source models.

Bio-SIEVE also outperforms traditional active-learning models that are specialised for an individual review. This demonstrates the promising capability of LLMs to assist in the SR process: a single model can be widely deployed for an entire SR domain, without the need for re-training per review task. Though we focused on biomedical applications of this technology, our models and training process could be applied to other domains such as software engineering or scientific systematic reviews. Our model is a first step towards this objective and we set a benchmark for generative language model solutions. 

The performance of Bio-SIEVE could be further improved by including few-shot prompting: both in training data as well as during inference. We did not experiment with few-shot prompts due to the limitations of model context window: the length of each sample makes it difficult to include additional examples in the prompt. However, using a mix of zero-shot and appropriately selected few-shot examples is likely to lead to improved performance, as discussed in \cite{longpreFlanCollectionDesigning2023}. 

Finally, we highlight the current shortcomings of our approach. Namely, that the exclusion reasons generated by our Bio-SIEVE-Multi variants were outperformed by ChatGPT. In our case, multi-task training was necessary to enable exclusion reasoning capability, but this worsened inclusion-exclusion performance. For future work, we plan to explore better methods of achieving multi-task capability in Bio-SIEVE, such as using a Mixture-of-Experts architecture \cite{DBLP:conf/iclr/ShazeerMMDLHD17}. We hope that adding a greater variety of tasks will eventually improve the model's reasoning capabilities and extend its functionality to form an effective generalised assistant for every stage of the SR process.

\section*{Acknowledgements}
We'd like to thank Ruth Wong at ScHARR for their input and facilitating the invaluable collaboration with ScHARR.

\bibliography{aaai24,custom}

\appendix

\section{Runtime Inference Analysis}
\label{appendix:runtime}

\begin{table}[h]
\centering
\begin{tabular}{|c|c|c|}
\hline
Batch Size & Memory Usage (GB) & Time/Sample (s) \\
\hline
1 & 12.8 & 1.39 \\
2 & 18.0 & 1.39 \\
3 & 23.1 & 1.13 \\
\hline
\end{tabular}
\caption{Metrics for runtime efficiency of our instruction pretrained, single task model with multiple batch sizes, averaged across 100 samples. All tests were carried out on the same RTX 3090.}
\label{table:runtime}
\end{table}

Bio-SIEVE is not useful if it is not easily accessible and efficient for reviewers to utilise. We ran performance analysis for our most promising model, the Guanaco7b Single variant, on an RTX 3090. As all models have the same number of parameters and architecture, this evaluation will transfer to all variants. Inference was carried out with 1 beam, no sampling and a temperature of 0 with each sample using the entire 2048 tokens in the LLaMA context length, therefore representing the worst case scenario. We measured maximum memory usage and time taken per sample for a batch size of 1, 2 and 3, averaging for each over 100 samples. Note that x-formers was also installed which improves performance of the attention mechanism. Results can be found in Table \ref{table:runtime}.

We find that the model performs with satisfactory speeds to apply on the scale of SR, with only minimal improvements with increased batch size. Memory usage shows that Bio-SIEVE will fit onto the memory of mid-end cards such as the RTX 2080ti but will need a reduced context length to ensure it does not go over 12GB memory usage. For higher-end consumer cards this should not become a problem.



\section{ChatGPT Prompts}
\label{appendix:chatgpt_prompt}

Refer to Figure \ref{figure:gpt_prompt} for the prompt used for querying ChatGPT for the inclusion/exclusion classification task. Also see Figure \ref{figure:gpt_prompt_reasons} for exclusion reasoning. Both prompts are adapted from \citet{syrianiAssessingAbilityChatGPT}.

\begin{figure}[H]
\noindent\fbox{%
    \parbox{\linewidth}{%
        I am screening papers for a systematic literature review.\newline
        The topic of the systematic review is \textbf{\{TOPIC\}}.\newline
        The objectives of the systematic review are \textbf{\{OBJECTIVES\}}\newline
        The selection criteria of the review is \textbf{\{SELECTION CRITERIA\}}\newline
        The study should focus exclusively on this topic.\newline
        Decide if the article should be included or excluded from the systematic review.\newline
        I give the title and abstract of the article as input.\newline
        Only answer Included or Excluded.\newline
        Be lenient. I prefer including papers by mistake rather than excluding them by mistake.\newline
        Title: \textbf{\{TITLE\}}\newline
        Abstract: \textbf{\{ABSTRACT\}}\newline
    }%
}
\caption{Prompt used to query ChatGPT as defined in \citet{syrianiAssessingAbilityChatGPT}. To fit the specificity of our task we also provide the additional information of the Objectives and Selection Criteria for the systematic review.} \label{figure:gpt_prompt}
\end{figure}

\begin{figure}[H]
\noindent\fbox{%
    \parbox{\linewidth}{
        I am screening papers for a systematic literature review.\newline
        The topic of the systematic review is \textbf{\{TOPIC\}}.\newline
        The objectives of the systematic review are \textbf{\{OBJECTIVES\}}\newline
        The selection criteria of the review is \textbf{\{SELECTION CRITERIA\}}\newline
        The study should focus exclusively on this topic and be within this selection criteria.\newline
        The following article has been excluded. Please provide the reason why it has been excluded as best you can.\newline
        I give the abstract of the article as input.\newline
        Only answer Included or Excluded.\newline
        Be concise and only provide a single reason.\newline
        Abstract: \textbf{\{ABSTRACT\}}\newline
    }%
}
\caption{Prompt used to query ChatGPT for exclusion reasoning.} \label{figure:gpt_prompt_reasons}
\end{figure}

\section{Data Preprocessing} \label{appendix:preprocessing}

Refer to Algorithm \ref{alg:preprocessing} for the strategy for tokenisation and preprocessing of prompts for the Instruct Cochrane dataset.

\begin{algorithm}[tb]
\caption{Instruct Cochrane Preprocessing}
\label{alg:preprocessing}
\textbf{Input}: Review with associated Included \& Excluded studies\\
\textbf{Parameter}: Max token length $m$, Prompt template token length $p$, Instruction template token length $l$, Tokeniser $T$\\
\textbf{Output}: list of Instructions, Inputs \& Outputs sets
\begin{algorithmic}[1] 
\STATE Set $S$ to EMPTY LIST
\STATE Assume valid Objectives $o$ \& Selection Criteria $s$
\FORALL{studies}
\STATE Assume valid Abstract $a$
\STATE Remove colons
\STATE Tokenise $o$, $s$ and $a$ with $T$
\STATE Concatenate $o$, $s$ and $a$ into $x$
\WHILE{$|x| + p + l > m$}
\STATE $z := \max(|o|, |s|, |a|)$
\IF{the last third of $z$ contains a full stop}
\STATE Truncate on full stop
\ELSE
\STATE Naively truncate
\ENDIF
\ENDWHILE
\STATE Construct input $i$ from prompt template
\STATE Construct task specific output $y$
\STATE Append instruction, $i$ and $y$ to $S$
\ENDFOR
\STATE \textbf{return} $S$
\end{algorithmic}
\end{algorithm}

\section{Prompt-Section Token Length Analysis}
\label{appendix:prompt_section_tokens}

See Table \ref{table:token_stats} for a detailed break down of the minimum, maximum and average token length for each different section relevant and obtainable from the cochrane library for each review. We ultimately utilised Objectives Short, Selection Criteria Short and Study Abstract but the analysis of other sections in the prompt could potentially improve performance, especially with greater crossover with PICOS. However, we chose not to use them as they were generally noisier and involved more truncation.

\begin{table*}[!bp]
\centering
\begin{tabular}{ |l|cccccccccc| }
\hline

\multicolumn{1}{|c|}{} &
    \multicolumn{1}{c}{} &
    \multicolumn{2}{|c}{\textbf{Objectives}} &
    \multicolumn{5}{|c}{\textbf{Selection Criteria}} & 
    \multicolumn{2}{|c|}{\textbf{Study}}  \\
\cline{3-11}
\multicolumn{1}{|c}{\multirow{-2}{*}} &
    \multicolumn{1}{|c}{\multirow{-2}{*}{\textbf{Title}}} &
    \multicolumn{1}{|c}{\textbf{Short}} &
    \multicolumn{1}{|c}{\textbf{Long}} &
    \multicolumn{1}{|c}{\textbf{Short}} &
    \multicolumn{1}{|c}{\textbf{Studies}} &
    \multicolumn{1}{|c}{\textbf{Pop.}} &
    \multicolumn{1}{|c}{\textbf{Intervention}} &
    \multicolumn{1}{|c}{\textbf{Outcome}} &
    \multicolumn{1}{|c}{\textbf{Title}} &
    \multicolumn{1}{|c|}{\textbf{Abstract}} \\
\cline{1-11}
\textit{Mean} &
    \multicolumn{1}{c|}{20.83} &
    \multicolumn{1}{c|}{56.41} &
    \multicolumn{1}{c|}{93.49} &
    \multicolumn{1}{c|}{80.20} &
    \multicolumn{1}{c|}{102.03} &
    \multicolumn{1}{c|}{145.75} &
    \multicolumn{1}{c|}{271.12} &
    \multicolumn{1}{c|}{224.12} &
    \multicolumn{1}{c|}{31.76} &
    \multicolumn{1}{c|}{447.21} \\
\textit{Max} &
    \multicolumn{1}{c|}{77} &
    \multicolumn{1}{c|}{481} &
    \multicolumn{1}{c|}{3,227} &
    \multicolumn{1}{c|}{394} &
    \multicolumn{1}{c|}{1,624} &
    \multicolumn{1}{c|}{2,474} &
    \multicolumn{1}{c|}{4,351} &
    \multicolumn{1}{c|}{9,956} &
    \multicolumn{1}{c|}{151} &
    \multicolumn{1}{c|}{68,340} \\
\textit{Min} &
    \multicolumn{1}{c|}{5} &
    \multicolumn{1}{c|}{13} &
    \multicolumn{1}{c|}{12} &
    \multicolumn{1}{c|}{8} &
    \multicolumn{1}{c|}{0} &
    \multicolumn{1}{c|}{0} &
    \multicolumn{1}{c|}{0} &
    \multicolumn{1}{c|}{0} &
    \multicolumn{1}{c|}{2} &
    \multicolumn{1}{c|}{1} \\
\cline{1-11}
\end{tabular}
\caption{Length of dataset fields according to tokenized length using the LLaMA tokenizer. The final prompt only utilised the short Objectives and Selection Criteria plus the study's Abstract. We leave utilisation of the other fields in the extension and alteration of the prompt to further work.}
\label{table:token_stats}
\end{table*}

\section{Detailed Topic Distributions}
\label{appendix:topic_dist}

Refer to Table~\ref{table:topic_dists} for an exact breakdown of the number of samples in each topic on the Cochrane Library for both the Train and Test splits. Table \ref{tab:reviews} shows the more fine-grained topic of each of the 13 reviews within Subset.

\begin{table*}[]
\centering
\begin{tabular}{ |l|cc| }
\hline
Topic & Train & Test \\
\hline
Child Health & 45488 & 267\\
Cancer & 10184 & 155 \\
Lungs \& Airways & 4968 & 93\\
Heart \& Circulation & 3915 & 74 \\
Infectious Disease & 3487 & 59 \\
Gynaecology & 2592 & 38 \\
Allergy \& Intolerance & 1792 & 13 \\
Tobacco, Drugs \& Alcohol & 1497 & 66\\
Genetic Disorders & 1401 & 44 \\
Endocrine \& Metabolic & 1145 & 6\\
Dentistry \& Oral Health & 1108 & 4 \\
Gastroenterology \& Hepatology & 1073 & 24 \\
Pain \& Anaesthesia & 785 & - \\
Mental Health & 759 & 8 \\
Effective Practice \& Health Systems & 728 & 20 \\
Pregnancy \& Childbirth & 699 & 5 \\
Consumer \& Communication Strategies & 680 & 19 \\
Developmental, Psychosocial \& Learning Problems & 634 & 35\\
Eyes \& Vision & 577 & 10 \\
Neurology & 507 & - \\
Ear, Nose \& Throat & 443 & 5\\
Complementary \& Alternative Medicine & 421 & 5\\
Neonatal Care & 328 & 1 \\
Insurance Medicine & 325 & 1\\
Orthopaedics \& Trauma & 261 & 24 \\
Rheumatology & 232 & 5 \\
Skin Disorders & 216 & 1\\
Urology & 199 & 5\\
Reproductive \& Sexual Health & 190 & - \\
Public Health & 178 & -\\
Kidney Disease & 160 & 1\\
Diagnosis & 140 & 15 \\
Wounds & 119 & 4 \\
Health \& Safety at Work & 82 & - \\
Health Professional Education & 40 & -\\
Methodology & 12 & -\\
Blood Disorders & 10 & -\\
\cline{1-3}
\end{tabular}
\caption{Number of samples for each topic in the train and test sets.}
\label{table:topic_dists}
\end{table*}

\begin{table*}[]
\centering
\begin{tabular}{|c|l|l|}
\hline
\textbf{No.} & \textbf{Specific Topic} & \textbf{DOI} \\
\hline
0 & Oral Health & \url{https://doi.org/10.1002/14651858.CD012213.pub2} \\
1 & Public Health & \url{https://doi.org/10.1002/14651858.CD011677.pub2} \\
2 & Developmental, Psychosocial and Learning Problems & \url{https://doi.org/10.1002/14651858.CD012955.pub2} \\
3 & Bone, Joint and Muscle Trauma & \url{https://doi.org/10.1002/14651858.CD012424.pub2} \\
4 & Urology & \url{https://doi.org/10.1002/14651858.CD011673.pub2} \\
5 & Developmental, Psychosocial and Learning Problems & \url{https://doi.org/10.1002/14651858.CD008524.pub4} \\
6 & Gynaecology and Fertility & \url{https://doi.org/10.1002/14651858.CD012165} \\
7 & Gynaecological, Neuro-oncology and Orphan Cancer & \url{https://doi.org/10.1002/14651858.CD013261.pub2} \\
8 & Haematology & \url{https://doi.org/10.1002/14651858.CD010981.pub2} \\
9 & Effective Practice and Organisation of Care & \url{https://doi.org/10.1002/14651858.CD009149.pub3} \\
10 & Drugs and Alcohol & \url{https://doi.org/10.1002/14651858.CD003020.pub3} \\
11 & Cystic Fibrosis and Genetic Disorders & \url{https://doi.org/10.1002/14651858.CD002008.pub5} \\
12 & Stroke/Heart & \url{https://doi.org/10.1002/14651858.CD013650.pub2} \\
\cline{1-3}
\end{tabular}
\caption{List of Reviews within the Review Subset with their Specific Topic areas and DOIs}
\label{tab:reviews}
\end{table*}

\section{Training Analysis}
\label{appendix:training}

\begin{figure*}[t]
\centering
\includegraphics[width=1\textwidth]{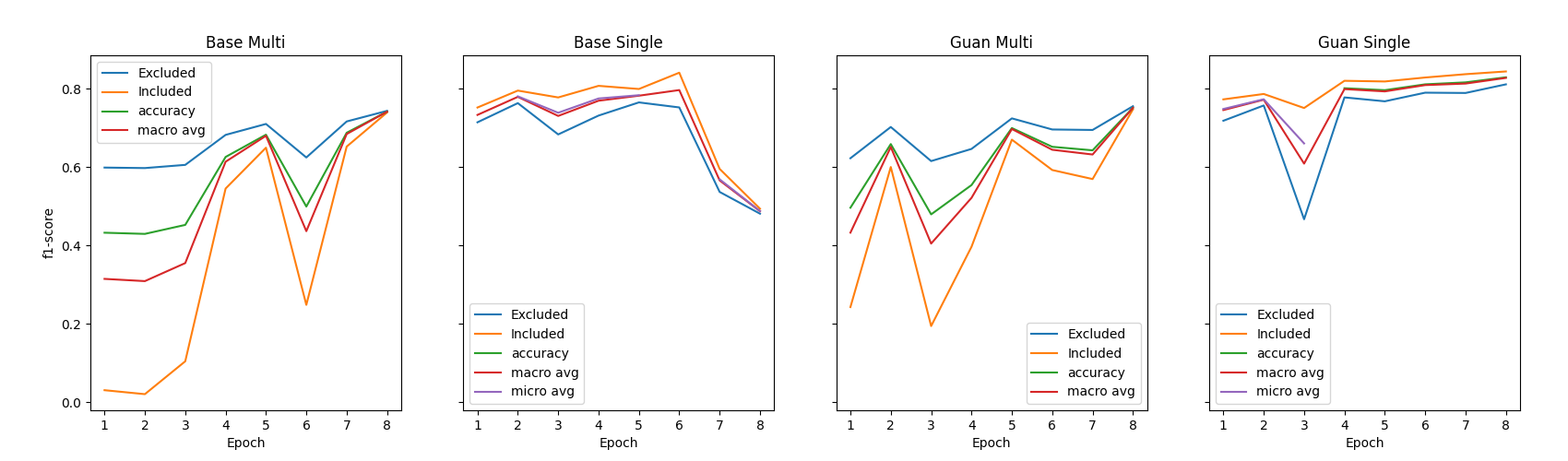} 
\caption{The f1-score performance for different metrics of the different model training regimes across epochs on the Test set.}
\label{topic_eval_epochs}
\end{figure*}

\begin{figure*}[t]
\centering
\includegraphics[width=1\textwidth]{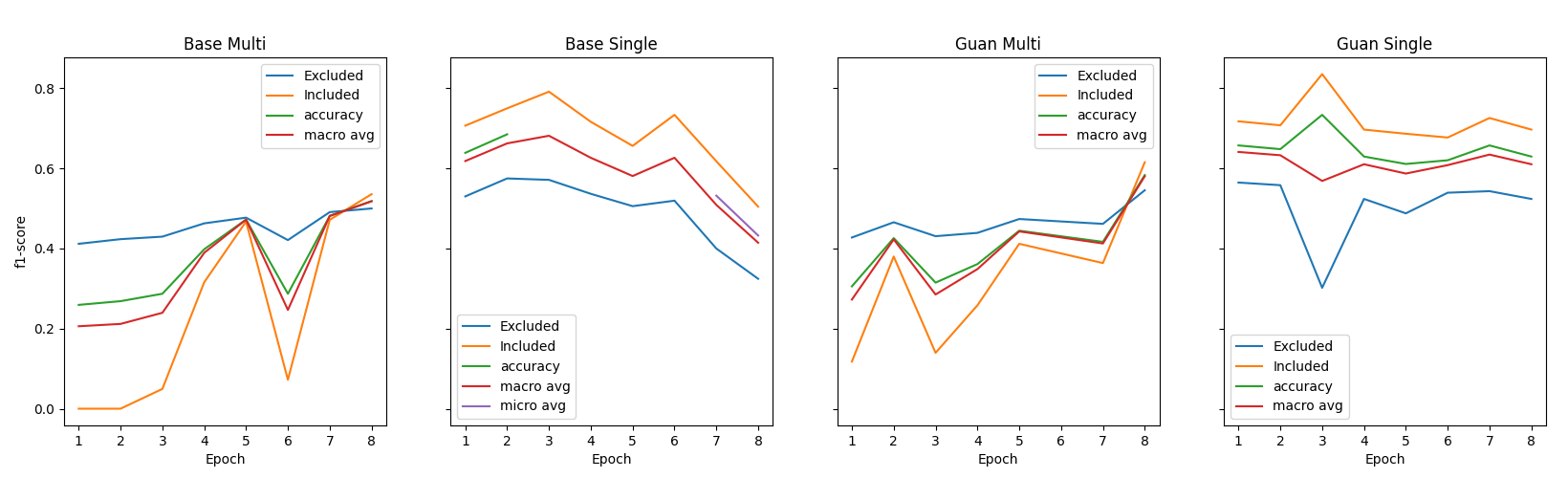} 
\caption{The f1-score performance for different metrics of the different model training regimes across epochs on the Safety-first set.}
\label{topic_gold_epochs}
\end{figure*}

\textbf{Training Analysis} Following the work in \citet{yangGoldilocksJustRightTuning2022}, we carried out an analysis of the models performance across epochs in order to select "just right" fine-tuning amount to sustain generalisation. Figure \ref{topic_eval_epochs} depicts the performance of each of the model variants over all 8 epochs for the Test set whilst Figure \ref{topic_gold_epochs} shows similar performance on the Safety-first set. Trends are consistent between sets but notable is the generally higher variance between include and exclude on the Safety-first set. Also the third epoch of the Guanaco7B single variant showing a sacrifice in Test set performance increasing leniency and improving Safety-first performance.

We used this method to pick our "just right" tuning for each model based on safety-first performance. This meant we chose epoch 8 for LLaMA7B Multi and Guanaco7B Multi, epoch 3 for LLaMA7B Single and epoch 7 for Guanaco7B Single. Interestingly we found that single task training overfit without instruction pretraining but did not with instruction pretraining. Additionally, multi-task training tended to result in larger fluctuation in performance between epochs which we hypothesis to be the model changing priority between tasks.

\end{document}